\documentclass[10pt,a4paper]{article}

\usepackage[margin=2.25cm]{geometry}
\usepackage[T1]{fontenc}
\usepackage[utf8]{inputenc}
\usepackage{lmodern}
\usepackage{microtype}
\usepackage{amsmath,amssymb}
\usepackage{booktabs}
\usepackage{array}
\usepackage{graphicx}
\usepackage{caption}
\usepackage{subcaption}
\usepackage{enumitem}
\usepackage{siunitx}
\usepackage{url}
\usepackage[hidelinks]{hyperref}
\usepackage{tikz}
\usetikzlibrary{arrows.meta,positioning,shapes.geometric}
\usepackage{float}
\usepackage{placeins}

\setlist{nosep}
\newcommand{\code}[1]{\texttt{#1}}

\title{\textbf{Wavelet-Based Parity Detection Revisited:}\\
Representation Dependence, Generalization, and Mechanistic Analysis}
\author{Ertu\u{g}rul Mutlu\\
RWTH Aachen University\\
Department of Computer Engineering\\
\texttt{ertugrul.mutlu@rwth-aachen.de}}
\date{Revised manuscript -- September 2026}

\begin{document}
\maketitle

\begin{abstract}
Parity is exactly determined by the least significant bit (LSB) of an integer, so machine learning is unnecessary for solving the task. This work instead uses parity as a deliberately over-engineered diagnostic for studying what a classical signal-processing pipeline preserves, amplifies, and loses when an exact symbolic property is embedded in a numerical representation. Integers from 0 to 10,000 are encoded as fixed-width 32-bit binary signals, decomposed with a level-3 Daubechies-2 (db2) discrete wavelet transform, summarized by the mean absolute coefficient magnitude, and clustered independently per subband with k-means. Importantly, the clustering itself is unsupervised, but cluster-to-parity mapping is calibrated using training labels only; a stratified 60/20/20 train/validation/test protocol prevents the label leakage present in an earlier version of the experiment.

After model selection on the validation split, the frozen configuration reaches \textbf{84.26\%} accuracy and balanced accuracy on the held-out test set (95\% Wilson CI: \textbf{82.60--85.79\%}). Across 20 stratified 80/20 resplits with different random seeds, the same configuration obtains \textbf{84.20\% $\pm$ 0.57\%} test accuracy. Ablations show that raw binary summary statistics reach only 61.30\%, while masking the LSB reduces performance to 48.15\%, consistent with chance. The level-3 approximation band (A3) alone retains 83.20\% validation accuracy, whereas the individual detail bands remain near chance. Performance is highly representation-dependent: moving the parity bit across the 32-bit signal changes validation accuracy from near-chance levels to 98.60\%, and changing the wavelet boundary mode ranges from 54.45\% to 83.20\%. Cross-magnitude extrapolation degrades from 79.98\% on 10,001--20,000 to 59.69\% on 100,001--1,000,000, although within-bit-length retraining remains around 78--88\%.

The results therefore do not support the interpretation that wavelets discover the arithmetic rule of parity. Rather, they show that parity-relevant information already present in the binary encoding can become more or less recoverable depending on multiscale filtering, spatial alignment, and boundary handling.
\end{abstract}


\section{Introduction}
Classification systems are often evaluated on difficult natural-data problems, but deliberately simple targets can provide unusually clean diagnostics of representation and inductive bias \cite{bishop2006,guyon2003}. Integer parity is an extreme example: in binary notation, the answer is exactly the least significant bit. The optimal symbolic solution therefore requires neither statistical learning nor signal processing.

That triviality is the point of this study. We ask a different question: \emph{if an exact symbolic property is routed through a lossy signal-processing and clustering pipeline, how much of the relevant information remains recoverable, and which representational choices control that recoverability?} Wavelet transforms are well suited to this diagnostic because they reorganize a signal into approximation and detail coefficients at multiple scales \cite{mallat1989,daubechies1992,addison2002}. K-means then provides a simple unsupervised partitioning mechanism whose behavior can be inspected directly \cite{macqueen1967,jain1999}.

An earlier version of this experiment reported approximately 69.67\% accuracy using wavelet-derived summary features and cluster-level oddness scores. Re-analysis exposed two methodological issues. First, the same labels used to compute cluster odd proportions were also used in evaluation, creating leakage. Second, the paper described the complete pipeline as unsupervised even though the cluster-to-parity mapping was label-dependent. The present revision rebuilds the evaluation protocol around a strict train/validation/test separation and reframes the task as \emph{unsupervised feature clustering with supervised cluster-to-class calibration}.

The revised study makes four contributions:
\begin{enumerate}[leftmargin=*]
    \item It establishes a leakage-safe primary result: a frozen db2, level-3, 32-bit model reaches 84.26\% accuracy on a held-out test set of 2,001 integers.
    \item It isolates the source of the signal through LSB masking, raw-feature baselines, subband ablation, simple coarse-filter baselines, spatial reordering, and boundary-mode controls.
    \item It evaluates stability through 20 repeated random splits and a learning curve extending to 100,001 integers.
    \item It separates in-range performance from cross-magnitude extrapolation and within-bit-length learnability, showing that the main failure mode is representation shift rather than an inability to model larger integers per se.
\end{enumerate}

The central conclusion is deliberately narrower than the original claim: the pipeline does not infer parity independently of the binary encoding. Instead, it exposes how a classical multiscale representation can preserve, amplify, or suppress access to symbolic information that is already present in the signal.

\section{Background and Related Work}
\subsection{Wavelet representations}
The discrete wavelet transform (DWT) decomposes a signal into low-frequency approximation coefficients and higher-frequency detail coefficients at progressively coarser scales. Multiresolution analysis provides a principled account of this decomposition \cite{mallat1989}, while Daubechies wavelets provide compactly supported filters with useful localization properties \cite{daubechies1992}. Wavelets have been widely used for denoising, compression, time-series analysis, and feature extraction \cite{addison2002}. In pattern-recognition pipelines, wavelet-derived statistics can serve as compact discriminative features; for example, wavelet features have been used in EEG classification \cite{subasi2007}.

The present work does not claim that parity resembles a conventional physical signal. Instead, it intentionally treats a fixed-width binary string as a one-dimensional signal so that the representation can be probed under controlled transformations.

\subsection{Clustering and feature engineering}
K-means partitions observations by minimizing within-cluster squared Euclidean distance \cite{macqueen1967}; its simplicity has made it a standard exploratory clustering method \cite{jain1999}. Here, each scalar wavelet summary is clustered independently. The clustering stage receives no parity labels. Labels enter only afterward, through a training-only mapping from cluster identity to the empirical fraction of odd integers in that cluster. This distinction is important: the method contains an unsupervised representation-partitioning step, but the resulting classifier is not fully unsupervised.

More broadly, the experiment belongs to feature engineering rather than competitive parity classification. Feature selection and representation design can strongly affect separability even when the underlying information content is unchanged \cite{guyon2003}. The controlled bit-position and boundary-mode experiments in this paper are designed to make that dependence explicit.

\subsection{Algorithmic and parity tasks as controlled diagnostics}
Small algorithmic problems are increasingly used as controlled settings for studying generalization, optimization, and learned representations. Power et al. used modular arithmetic tasks to demonstrate \emph{grokking}, where test generalization can emerge long after training-set memorization \cite{power2022}. Nanda et al. subsequently reverse-engineered a transformer trained on modular addition and showed that its solution is organized around Fourier structure, illustrating how simple arithmetic tasks can support unusually detailed mechanistic analysis \cite{nanda2023}.

Parity has an especially long-standing role as a diagnostic Boolean function. Daniely and Malach showed that, under specific distributions, depth-two neural networks trained with gradient methods can learn sparse parity functions in regimes where fixed linear representations are inefficient \cite{daniely2020}. Barak et al. later studied sparse parity learning as a probe of training dynamics and found that SGD can make gradual, initially hidden progress toward the relevant parity structure before abrupt improvements become visible in error metrics \cite{barak2022}. These studies ask how adaptive neural learners acquire algorithmic or parity rules. The present work addresses a different but complementary question: given a fixed, non-neural signal transform and a target bit already present in the input, which representation choices make that symbolic information accessible to simple clustering?

\section{Methodology}
\subsection{Binary signal representation}
For each non-negative integer $n$, we construct a fixed-width 32-bit vector
\begin{equation}
    \mathbf{x}(n) = [b_{31}, b_{30}, \ldots, b_1, b_0] \in \{0,1\}^{32},
\end{equation}
where $b_0$ is the least significant bit. Numbers are left-zero-padded, so $n=5$ becomes
\begin{equation*}
    00000000000000000000000000000101.
\end{equation*}
Parity is exactly $y(n)=b_0$. This means the target information is explicitly present in the representation; the purpose of the experiment is to measure how recoverable that information remains after transformation and summary.

The fixed width is a design choice, not a requirement of the DWT. A 32-bit width was selected because it supports level-3 db2 decomposition while avoiding the unnecessary padding of 64- or 128-bit alternatives, whose validation accuracies were nearly identical.

\subsection{Wavelet decomposition and feature extraction}
The primary representation applies a level-3 db2 DWT with symmetric boundary extension. PyWavelets returns
\begin{equation}
    \mathcal{C}(n)=\{\mathbf{c}_{A3}(n),\mathbf{c}_{D3}(n),\mathbf{c}_{D2}(n),\mathbf{c}_{D1}(n)\}.
\end{equation}
For each subband $b$, the selected feature is the mean absolute coefficient magnitude (MAV):
\begin{equation}
    m_b(n) = \frac{1}{|\mathbf{c}_b(n)|}\sum_i |c_{b,i}(n)|.
    \label{eq:mav}
\end{equation}
Energy and coefficient standard deviation were also evaluated during validation, but MAV alone gave the best validation accuracy while yielding the simplest final representation.

\subsection{Unsupervised clustering with supervised calibration}
For each subband $b\in\{A3,D3,D2,D1\}$, a one-dimensional k-means model with $k=2$, \code{n\_init=20}, and random seed 42 is fit \emph{only} on the training MAV values. Let $g_b(n)\in\{0,1\}$ denote the resulting cluster assignment. We then compute a training-only empirical odd fraction
\begin{equation}
    q_{b,c}=
    \frac{\sum_{n\in\mathcal{T}} \mathbb{1}[g_b(n)=c]y(n)}
         {\sum_{n\in\mathcal{T}} \mathbb{1}[g_b(n)=c]},
    \label{eq:calibration}
\end{equation}
for cluster $c$. These quantities are used as \emph{oddness scores}; we do not interpret them as calibrated predictive probabilities.

For a new integer $n$, the final score averages the four subband contributions:
\begin{equation}
    S(n)=\frac{1}{4}\sum_{b\in\{A3,D3,D2,D1\}} q_{b,g_b(n)}.
    \label{eq:score}
\end{equation}
The prediction rule is
\begin{equation}
    \hat{y}(n)=
    \begin{cases}
      1 & S(n)>0.5,\\
      0 & \text{otherwise.}
    \end{cases}
\end{equation}

\begin{figure}[H]
\centering
\begin{tikzpicture}[
  node distance=6mm and 8mm,
  box/.style={draw, rounded corners, align=center, minimum height=9mm, text width=30mm, font=\small},
  arrow/.style={-{Latex[length=2mm]}, thick}
]
\node[box] (n) {Integer $n$\\$0\ldots 10{,}000$};
\node[box, right=of n] (bits) {32-bit binary\\left zero-padding};
\node[box, right=of bits] (dwt) {db2 DWT\\level 3, symmetric};
\node[box, below=of dwt] (feat) {MAV per subband\\$A3,D3,D2,D1$};
\node[box, left=of feat] (km) {$k$-means per subband\\$k=2$};
\node[box, left=of km] (cal) {Train-label calibration\\$q_{b,c}$};
\node[box, below=of cal] (score) {Mean oddness score\\$S(n)$};
\node[box, right=of score] (pred) {Prediction\\$S(n)>0.5$};
\draw[arrow] (n)--(bits);
\draw[arrow] (bits)--(dwt);
\draw[arrow] (dwt)--(feat);
\draw[arrow] (feat)--(km);
\draw[arrow] (km)--(cal);
\draw[arrow] (cal)--(score);
\draw[arrow] (score)--(pred);
\end{tikzpicture}
\caption{Frozen primary pipeline. K-means fitting is unsupervised, whereas cluster-to-parity calibration uses training labels only.}
\label{fig:pipeline}
\end{figure}

\subsection{Data split, model selection, and final test}
The primary dataset contains all 10,001 integers from 0 through 10,000. A stratified split with random seed 42 produces 6,000 training samples, 2,000 validation samples, and 2,001 held-out test samples. Hyperparameters and representation choices are selected using training and validation data only. After freezing the configuration, the final k-means models and cluster calibration values are re-fit on the combined 8,000 train+validation samples and evaluated once on the 2,001-sample test set.

Table~\ref{tab:selection} summarizes the main validation decisions. Wavelet-family comparison was performed at 64 bits and level 3. db2 and sym2 tied at 82.90\%; db2 was retained as the canonical choice. Increasing the decomposition depth from level 1 to 3 substantially improved validation accuracy. Once level 3 was fixed, 32-, 64-, and 128-bit representations differed by at most 0.50 percentage points; 32 bits was retained for parsimony. Finally, MAV alone outperformed the larger three-statistic feature set on the 32-bit configuration.

\begin{table}[H]
\centering
\caption{Validation-stage model selection. Results belong to the stated comparison settings and were not obtained by evaluating the held-out primary test set.}
\label{tab:selection}
\small
\begin{tabular}{lllS[table-format=2.2]}
\toprule
Selection question & Candidate & Fixed context & {Val. acc. (\%)}\\
\midrule
Wavelet & Haar & 64-bit, L3 & 61.50\\
        & db2 & 64-bit, L3 & 82.90\\
        & sym2 & 64-bit, L3 & 82.90\\
        & coif1 & 64-bit, L3 & 80.90\\
\midrule
Depth & L1 & db2, 64-bit & 63.45\\
      & L2 & db2, 64-bit & 75.80\\
      & L3 & db2, 64-bit & 82.90\\
\midrule
Width & 32 & db2, L3 & 82.50\\
      & 64 & db2, L3 & 82.90\\
      & 128 & db2, L3 & 83.00\\
\midrule
Feature & Energy & db2, L3, 32-bit & 80.40\\
        & MAV & db2, L3, 32-bit & 83.20\\
        & Std. dev. & db2, L3, 32-bit & 81.30\\
        & Energy+MAV+Std. & db2, L3, 32-bit & 82.50\\
\bottomrule
\end{tabular}
\end{table}

\subsection{Evaluation metrics}
Accuracy is the primary metric. Because every controlled split is parity-balanced or stratified, balanced accuracy is also reported; in the primary test it is essentially identical to accuracy. We report 95\% Wilson score intervals for binomial accuracy. Confusion-matrix counts are retained for error analysis. No claim of statistical significance is made for small validation differences among neighboring configurations. For a small number of paired validation ablations evaluated on the same 2,000 examples, we additionally report exact two-sided McNemar tests to distinguish large representation effects from negligible feature-set differences; these secondary tests do not affect model selection or the held-out primary result.

\section{Experimental Analyses}
The revised study separates the primary evaluation from secondary analyses. The original 2,001-sample primary test is not used for V3/V5 validation ablations or V4 out-of-distribution fitting. Secondary experiments probe representation dependence rather than retroactively optimizing the frozen primary model.

\subsection{Core baselines and ablations}
We compare the selected wavelet representation against: (i) identical summary statistics applied directly to the raw 32-bit vector, (ii) right rather than left padding, (iii) masking the natural LSB before feature extraction, (iv) a 50\% majority baseline, and (v) an exact LSB reference that is not a machine-learning model. For the right-padding and LSB-masking ablations, all other selected settings are held fixed, including the MAV-only feature representation.

\subsection{Subband and coarse-representation controls}
To determine where the discriminative information resides, the four DWT subbands are evaluated individually and in combinations. We also compare db2-A3 against Haar-A3, direct raw MAV, three-level average pooling, and a generic triangular low-pass/downsampling baseline. Each coarse baseline contributes one scalar feature per integer and uses the same $k=2$ clustering/calibration protocol.

\subsection{Robustness, scaling, and generalization}
Three complementary experiments are used.
\begin{enumerate}[leftmargin=*]
    \item \textbf{Repeated splits:} the frozen representation is evaluated across 20 fresh stratified 80/20 train/test splits with seeds 0--19.
    \item \textbf{Learning curve:} on integers 0--100,000, a fixed 20\% holdout is evaluated while training subsets increase from 500 to 80,000 samples.
    \item \textbf{Cross-magnitude OOD:} a model fit and calibrated on the original 8,000 train+validation integers from 0--10,000 is applied without recalibration to 10,001--20,000, 20,001--50,000, 50,001--100,000, and 100,001--1,000,000.
\end{enumerate}
A separate bit-length-controlled experiment retrains within 10-, 12-, 14-, 16-, 18-, and 20-bit magnitude bands to distinguish cross-range distribution shift from intrinsic difficulty at larger magnitudes. For the 18- and 20-bit bands, computation is capped at a stratified sample of 50,000 integers per band using deterministic seeds derived from the global random seed; each sampled band is then split 80/20 into train and test sets. Smaller bands are evaluated exhaustively before the same 80/20 split.

\subsection{Mechanistic controls}
The same 32 bits are rearranged without adding or deleting information. In the LSB-position sweep, the parity bit is swapped with each of the 32 signal positions. Additional controls reverse all bits or apply one fixed random permutation (seed 42). Finally, DWT boundary modes (symmetric, reflect, zero, constant, and periodization) and k-means cluster counts $k\in\{2,4,8,16\}$ are varied on the validation split. These are explanatory sensitivity analyses; they do not replace the frozen primary configuration.

\section{Results}
\subsection{Primary held-out performance}
The frozen model achieves \textbf{84.26\%} accuracy on 2,001 previously held-out integers, with a 95\% Wilson confidence interval of \textbf{82.60--85.79\%}. Balanced accuracy is also 84.26\%. The confusion matrix is $TN=837$, $FP=164$, $FN=151$, and $TP=849$, corresponding to precision 83.81\%, recall 84.90\%, specificity 83.62\%, and F1 84.35\%.

\begin{figure}[H]
\centering
\includegraphics[width=0.92\linewidth]{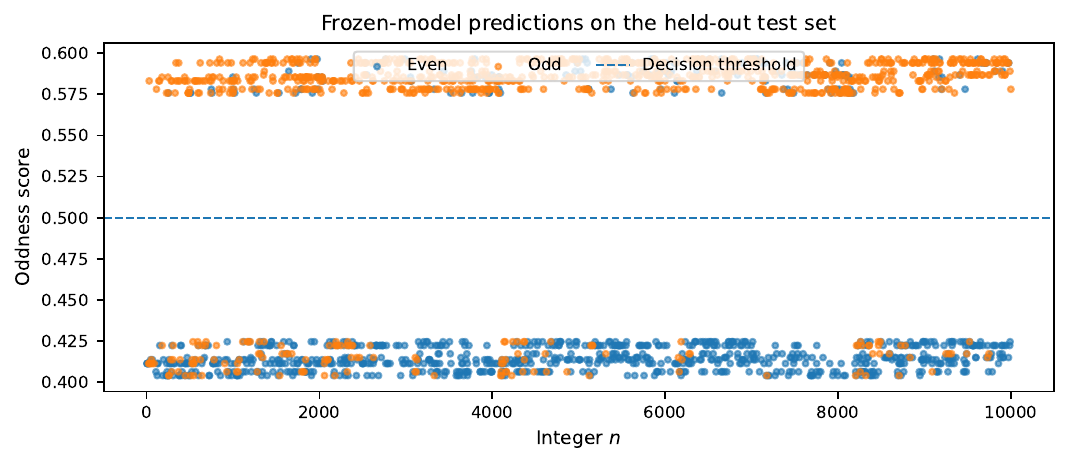}
\caption{Oddness scores on the frozen held-out test set. The test set was not used for model or representation selection.}
\label{fig:test-scores}
\end{figure}

\subsection{The signal depends on the explicit parity bit and the representation}
\begin{figure}[H]
\centering
\includegraphics[width=0.92\linewidth]{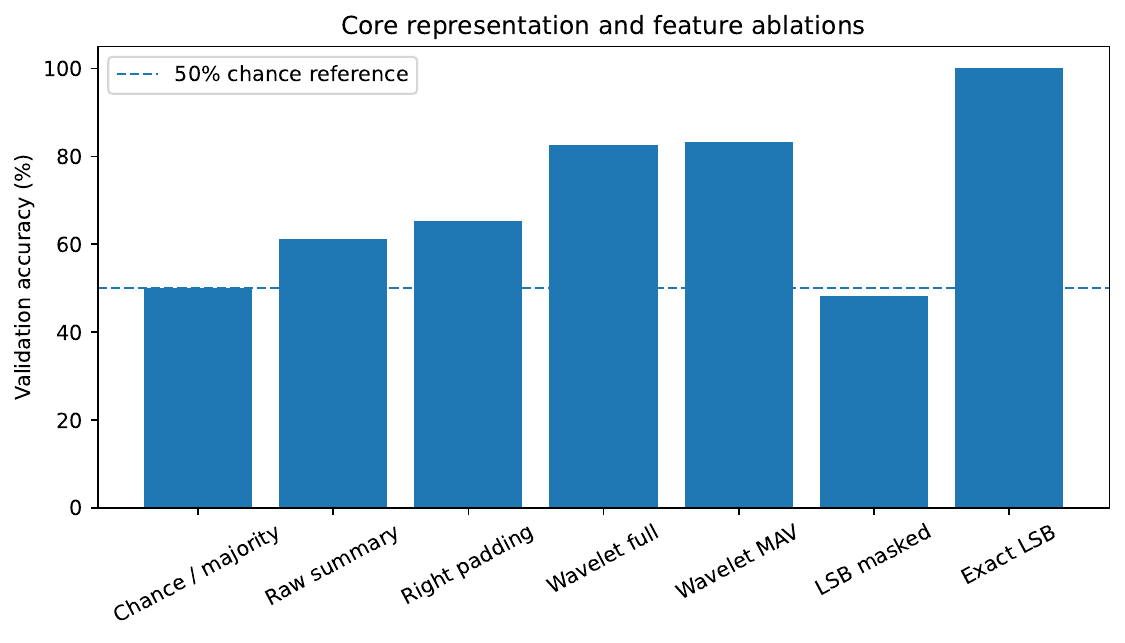}
\caption{Core validation ablations. The exact-LSB rule is an arithmetic reference, not an ML model. The primary held-out test remains separate from this validation analysis.}
\label{fig:core-ablation}
\end{figure}

Figure~\ref{fig:core-ablation} shows the central ablations. Applying the summary features directly to raw binary vectors reaches 61.30\% validation accuracy. The selected wavelet MAV representation reaches 83.20\%. Masking the natural LSB before wavelet extraction reduces accuracy to 48.15\% (95\% CI: 45.97--50.34\%), which is consistent with chance. This result is decisive for interpretation: the pipeline is not reconstructing parity after the explicit parity bit has been removed.

Right padding also reduces performance to 65.20\%, demonstrating that the location of the binary pattern within the fixed signal strongly affects separability.

Because these validation predictions are paired example-by-example, exact McNemar tests provide an additional check on the magnitude of the central ablations. Comparing wavelet MAV with raw MAV yields 438 discordant examples on which the wavelet model is correct and the raw model is wrong, versus 0 in the opposite direction ($p=2.82\times10^{-132}$). Comparing wavelet MAV with the MAV-only LSB-masked condition yields 747 versus 46 discordant examples ($p=4.57\times10^{-164}$). Comparing wavelet MAV with the MAV-only right-padding condition yields 562 versus 202 discordant examples ($p=4.69\times10^{-40}$). In contrast, the small difference between MAV alone and the full Energy+MAV+Std. feature set is not statistically distinguishable on this split (55 versus 41 discordant examples, $p=0.184$). We therefore retain MAV alone for parsimony rather than claiming a meaningful accuracy advantage over the full feature set.

\subsection{Discriminative information is concentrated in A3}
The level-3 approximation band $A3$ alone reaches 83.20\% validation accuracy, exactly matching the all-band prediction in this split. In contrast, $D3$, $D2$, and $D1$ individually achieve 52.35\%, 50.30\%, and 50.40\%, respectively. Combinations containing only detail bands likewise remain near chance.

\begin{figure}[H]
\centering
\includegraphics[width=0.78\linewidth]{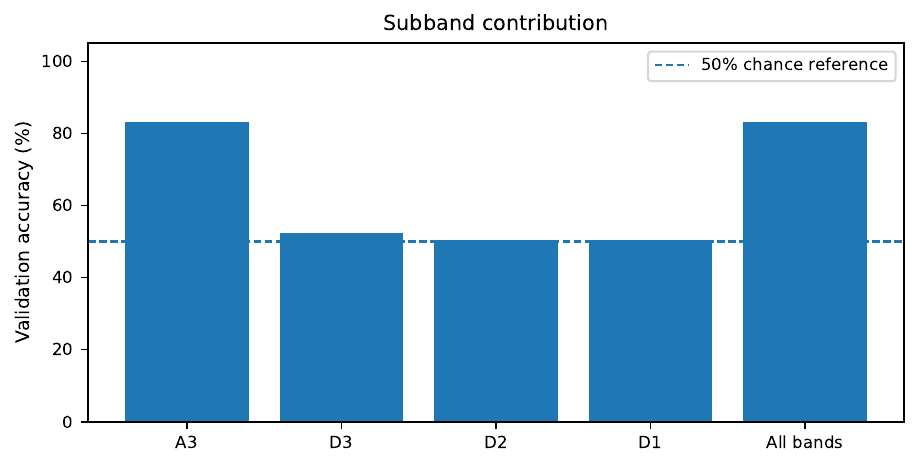}
\caption{Subband ablation. Nearly all usable discriminative signal is retained by the level-3 approximation band, while the detail bands are individually close to chance.}
\label{fig:subbands}
\end{figure}

This finding is initially counter-intuitive because parity is determined by a single fine-scale bit. The coarse-baseline control clarifies that the effect is not explained by generic averaging alone. db2-A3 MAV obtains 83.20\%, whereas Haar-A3, raw MAV, and three-level average pooling each obtain 61.30\%, and the triangular low-pass baseline obtains 51.40\%.

\begin{figure}[H]
\centering
\includegraphics[width=0.82\linewidth]{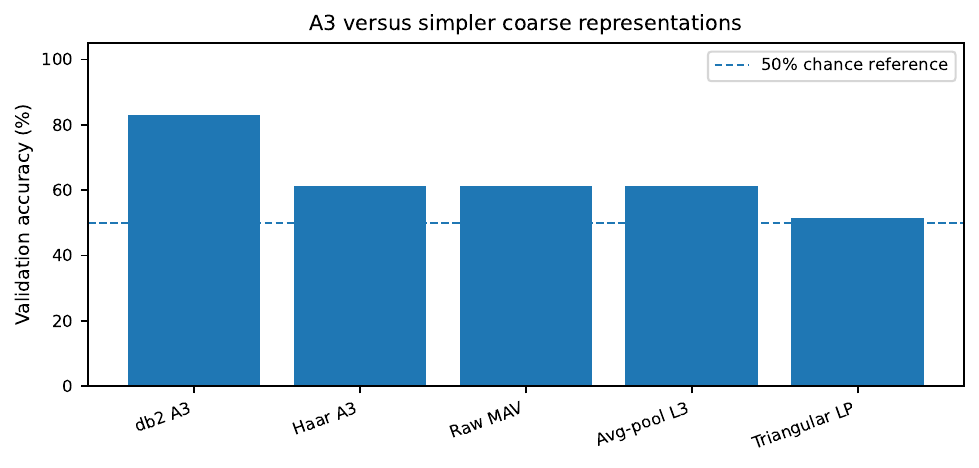}
\caption{db2-A3 compared with simpler coarse representations. The db2 approximation feature is substantially more discriminative than direct averaging or the tested non-wavelet low-pass controls.}
\label{fig:coarse}
\end{figure}

\subsection{The primary result is stable across random splits}
Across 20 stratified 80/20 splits, the frozen configuration obtains a mean accuracy of \textbf{84.20\% $\pm$ 0.57\% SD}, with a range of 83.06--85.21\%. This indicates that the 84.26\% primary result is not an isolated favorable split. Because test subsets overlap across seeds, the standard deviation is descriptive rather than an estimate from independent datasets.

\begin{figure}[H]
\centering
\includegraphics[width=0.78\linewidth]{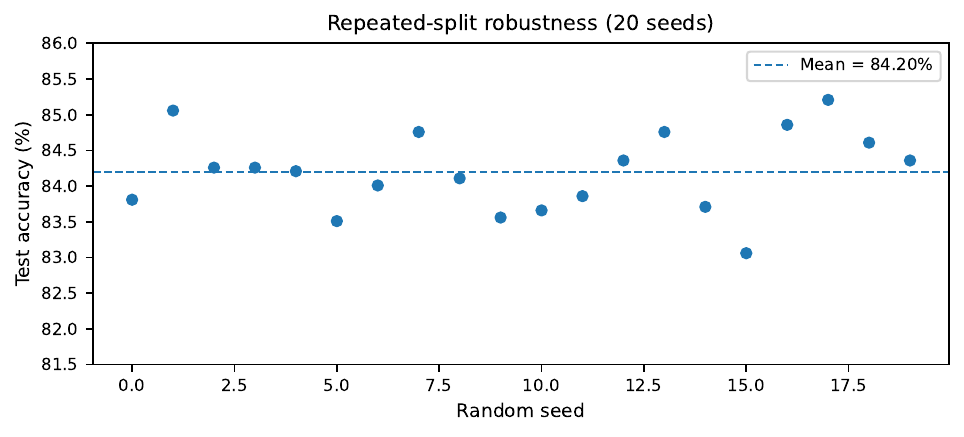}
\caption{Repeated-split robustness for 20 random seeds.}
\label{fig:seeds}
\end{figure}

\subsection{More training data does not remove the representation ceiling}
On the broader 0--100,000 distribution, increasing the training subset from 500 to 80,000 samples changes fixed-holdout accuracy only from 79.47\% to 79.72\%. The curve saturates rapidly: 2,500 samples already yield 79.68\%. Thus, within this model class and representation, the principal bottleneck is not the amount of calibration data.

\begin{figure}[H]
\centering
\includegraphics[width=0.78\linewidth]{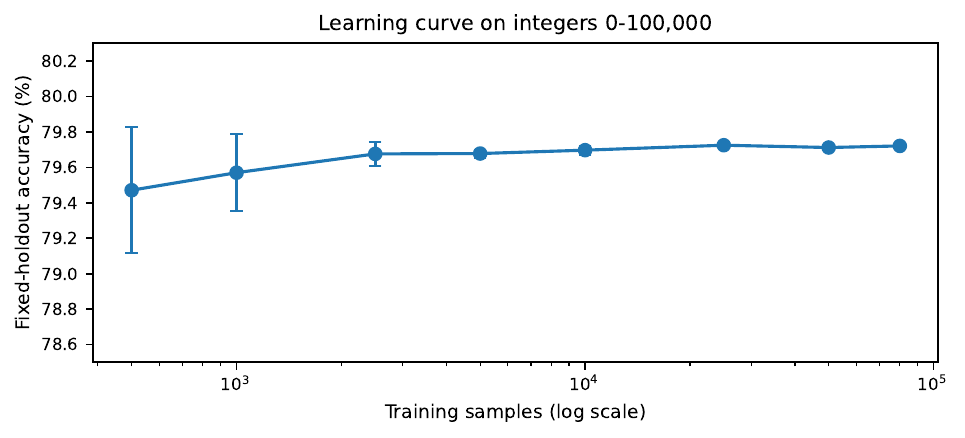}
\caption{Learning curve on integers 0--100,000 with a fixed 20\% holdout. The wider magnitude distribution is different from the primary 0--10,000 experiment, so the vertical level should not be interpreted as a replacement primary score.}
\label{fig:learning}
\end{figure}

\subsection{Cross-magnitude extrapolation degrades systematically}
The frozen model fit on the original 8,000 train+validation samples generalizes above chance to larger magnitudes but degrades monotonically with distance from the training range: 79.98\% on 10,001--20,000, 71.51\% on 20,001--50,000, 66.32\% on 50,001--100,000, and 59.69\% on 100,001--1,000,000. The far-OOD errors are strongly asymmetric: among 900,000 samples in the largest range, the model produces 360,567 false positives but only 2,217 false negatives, indicating a substantial shift in the cluster geometry relative to the training calibration.

\begin{figure}[H]
\centering
\includegraphics[width=0.82\linewidth]{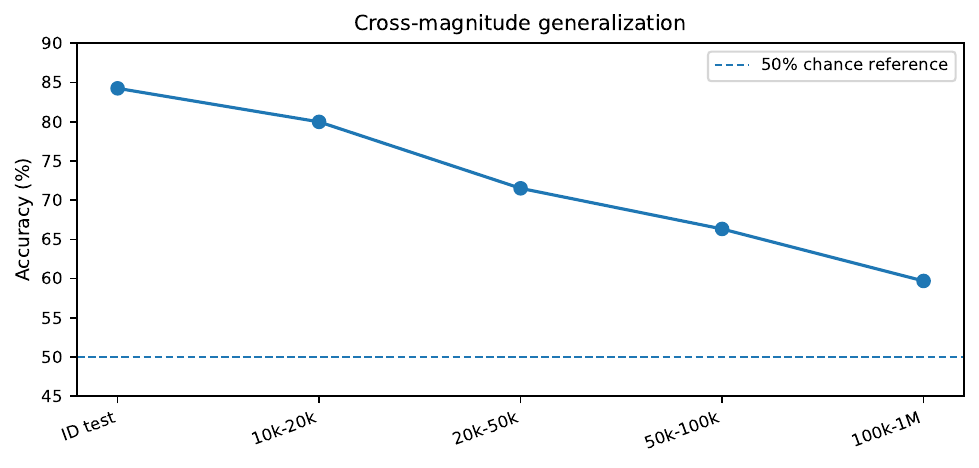}
\caption{Cross-magnitude generalization without recalibration. The first point is the frozen primary held-out test; subsequent points use progressively larger OOD integer ranges.}
\label{fig:ood}
\end{figure}

Within-band retraining paints a different picture. When separate models are trained and tested within fixed bit-length bands, test accuracy remains 78.64\% (10 bit), 87.56\% (12 bit), 83.59\% (14 bit), 80.50\% (16 bit), 78.38\% (18 bit), and 79.29\% (20 bit). The 10-bit estimate has a wide interval because only 103 test samples are available. These results suggest that the main OOD failure is cross-magnitude representation shift rather than a complete loss of learnability at larger bit lengths.

\subsection{Spatial location and boundary handling dominate the mechanism}
The LSB-position sweep reveals extreme sensitivity to where the parity-carrying bit appears in the 32-bit signal. With the parity bit in its natural position, validation accuracy is 83.20\%. Moving it through many interior positions yields roughly 53--75\%, while swapping it into positions 30 and 31 from the LSB side yields 93.85\% and 98.60\%, respectively. Since each operation is a pure bit permutation, information content is preserved; only spatial arrangement changes.

\begin{figure}[H]
\centering
\includegraphics[width=0.92\linewidth]{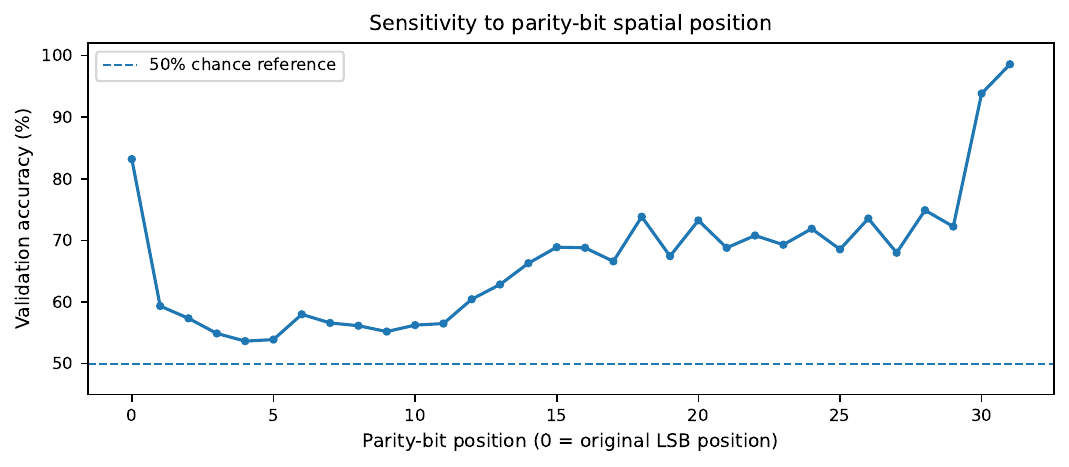}
\caption{Validation accuracy as the original parity bit is swapped across the 32-bit signal. Position 0 is the natural LSB position; position 31 moves the parity bit to the MSB-side column.}
\label{fig:position}
\end{figure}

The broader reordering controls are consistent with this conclusion: the original ordering yields 83.20\%, full reversal yields 88.75\%, and one fixed random permutation yields 91.00\%. The random-permutation result is a single control condition and should not be interpreted as evidence that random permutations generally improve performance.

Boundary handling is similarly influential. With all other settings frozen, symmetric extension yields 83.20\%, constant extension 77.20\%, zero extension 71.80\%, periodization 64.85\%, and reflect extension 54.45\%. Consequently, the observed separation is best understood as an interaction among the binary layout, db2 filters, multilevel decomposition, and boundary treatment rather than a wavelet-family effect in isolation.

\begin{figure}[H]
\centering
\includegraphics[width=0.78\linewidth]{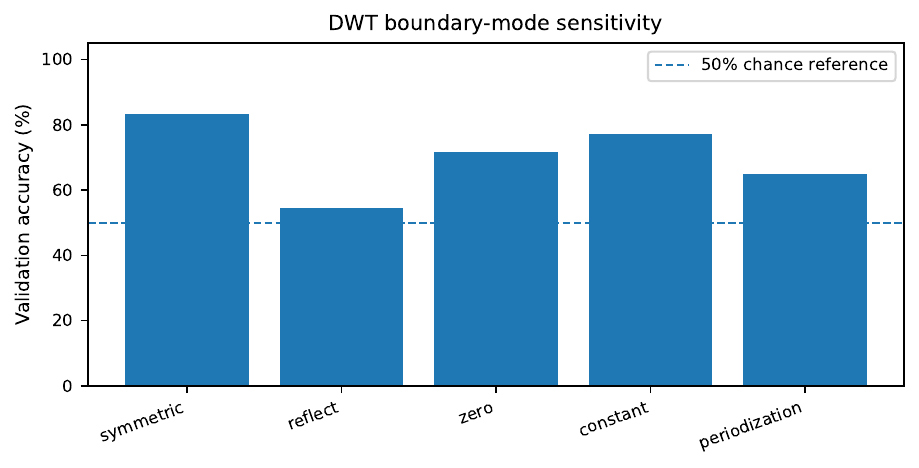}
\caption{Validation accuracy under different PyWavelets boundary-extension modes with the rest of the configuration frozen.}
\label{fig:boundary}
\end{figure}

Finally, varying the k-means cluster count has only a small effect: $k=2,4,8,16$ yields 83.20\%, 83.25\%, 83.40\%, and 83.70\% validation accuracy, respectively. The range is less than one percentage point, so the qualitative conclusions are not tied to choosing exactly two clusters. The original $k=2$ configuration remains frozen for the primary result.

\section{Discussion}
\subsection{What is the model actually learning?}
The combined ablations sharply constrain the interpretation. First, parity is explicitly encoded in $b_0$. Second, masking $b_0$ reduces performance to chance. Third, merely rearranging the same bits can push validation accuracy from near chance to almost perfect separation. These observations are inconsistent with a claim that the pipeline discovers the abstract arithmetic rule of parity independently of representation.

A more accurate interpretation is that the DWT and the subsequent scalar summaries create a geometry in which the already-present parity bit becomes more or less accessible to simple clustering. The method is therefore a controlled example of \emph{representation-mediated recoverability}. The distinction matters because an exact symbolic rule and a statistical proxy for information carried by that rule are fundamentally different objects.

\subsection{Why does A3 dominate?}
The A3 result is particularly informative. A naive intuition might predict that a one-bit property would appear primarily in the finest-scale detail band. Instead, the detail bands are near chance while the approximation band retains almost the entire predictive signal. The coarse-baseline controls show that this is not equivalent to counting ones or performing ordinary average pooling: db2-A3 is about 22 percentage points better than raw MAV and average pooling on the validation split.

The strong dependence on spatial position and boundary extension suggests that repeated db2 filtering and downsampling combine the parity bit with neighboring and boundary values in a location-specific way. The resulting A3 magnitude is therefore a structured proxy for parity under this particular fixed-width encoding. Establishing a closed-form characterization of that proxy would be an interesting mathematical extension, but it is not required for the empirical conclusion reached here.

\subsection{Generalization is limited by representation shift}
The OOD results expose an important limitation. A cluster geometry calibrated on 0--10,000 does not remain invariant as the magnitude distribution changes; accuracy approaches chance on the broad 100,001--1,000,000 range. Yet fresh within-band models still reach roughly 78--88\%, showing that larger magnitudes are not intrinsically inaccessible to the representation. The problem is extrapolation across regimes, not a simple monotonic loss of signal with integer size.

The learning curve reinforces this interpretation. On a fixed 0--100,000 distribution, hundreds of training samples are already enough to approach the eventual plateau. Adding tens of thousands of samples does not materially raise the ceiling. The limiting factor is therefore representational stability and the restricted one-dimensional cluster model, not sample scarcity.

\subsection{Limitations}
Several limitations should be kept explicit.
\begin{itemize}[leftmargin=*]
    \item \textbf{The task is intentionally artificial.} Parity has an exact one-bit solution, and this approach is computationally inferior to direct bit inspection.
    \item \textbf{The classifier is not fully unsupervised.} K-means fitting is label-free, but cluster-to-class calibration uses training labels.
    \item \textbf{The target is present in the input.} The experiments study preservation and exposure of known symbolic information, not discovery of parity from a representation that omits the parity bit.
    \item \textbf{Model selection used one validation split.} Repeated-seed analysis evaluates the frozen configuration and supports stability, but it is not nested cross-validation of the full model-selection process.
    \item \textbf{Hyperparameter selection was staged rather than jointly exhaustive.} Wavelet family, decomposition depth, signal width, and feature set were evaluated in successive controlled comparisons rather than through a complete Cartesian search over all combinations. Table~\ref{tab:selection} therefore documents a practical staged selection procedure, not a proof that the frozen configuration is globally optimal over the full design space.
    \item \textbf{Mechanistic controls are diagnostic, not new primary models.} The 98.60\% LSB-position result and 91.00\% fixed-permutation result deliberately alter spatial layout after the original test protocol was frozen; they must not replace the 84.26\% held-out result.
    \item \textbf{Boundary and layout sensitivity is substantial.} This is a scientific finding of the study, but it also limits claims of representation-invariant generalization.
    \item \textbf{The analysis is empirical.} The present work does not provide a formal proof explaining why db2-A3 with symmetric extension induces the observed cluster geometry.
\end{itemize}

\section{Conclusion}
This paper revisits a deliberately over-engineered parity classifier and turns it into a controlled study of feature representation. A leakage-safe db2 wavelet pipeline with training-only cluster calibration reaches 84.26\% accuracy on a held-out test set, with stable performance across repeated random splits. However, the stronger contribution is explanatory rather than predictive. Removing the LSB destroys performance; A3 contains nearly all useful signal; simple coarse averages do not reproduce the db2 result; spatial reordering and boundary extension dramatically alter accuracy; and cross-magnitude extrapolation degrades even when within-band learnability remains high.

The evidence therefore supports a precise conclusion: classical signal-processing features can make an explicitly encoded symbolic property statistically accessible to simple clustering, but the effect is highly dependent on how that property is arranged in the signal. In this setting, wavelets do not replace symbolic reasoning. They provide a useful microscope for studying how symbolic information is reshaped by numerical representation.

\section*{Reproducibility Note}
Code and experiment artifacts for this manuscript are available in the
\href{https://github.com/Ertugrulmutlu/Using-Wavelets-and-Clustering-to-Predict-Odd-or-Even-Numbers}{project's GitHub repository}.
The manuscript release is identified by the immutable Git tag \code{paper-v2}. The tagged bundle contains the V2--V5 Python scripts, frozen split indices, JSON/CSV metrics, NPZ prediction outputs, and generated figures used for this manuscript. The primary result is defined by the V3 frozen configuration; V4 and V5 experiments are secondary robustness and mechanism analyses and do not replace the held-out primary test. Readers reproducing the arXiv version should use the tagged release rather than the repository's moving default branch.

\appendix
\section{Additional Secondary Results}

\subsection{Bit-length-controlled results}
\begin{table}[H]
\centering
\caption{Within-band train/test experiments. Each bit-length band receives its own stratified 80/20 split and calibration; the 18- and 20-bit bands use the 50,000-sample stratified cap described in Section~4.3.}
\small
\begin{tabular}{rrrrS[table-format=2.2]l}
\toprule
Bits & Range start & Range end & Test $N$ & {Test acc. (\%)} & 95\% CI\\
\midrule
10 & 512 & 1023 & 103 & 78.64 & 69.77--85.45\\
12 & 2048 & 4095 & 410 & 87.56 & 84.01--90.41\\
14 & 8192 & 16383 & 1639 & 83.59 & 81.72--85.30\\
16 & 32768 & 65535 & 6554 & 80.50 & 79.52--81.44\\
18 & 131072 & 262143 & 10000 & 78.38 & 77.56--79.18\\
20 & 524288 & 1048575 & 10000 & 79.29 & 78.48--80.07\\
\bottomrule
\end{tabular}
\end{table}

\begin{figure}[H]
\centering
\includegraphics[width=0.75\linewidth]{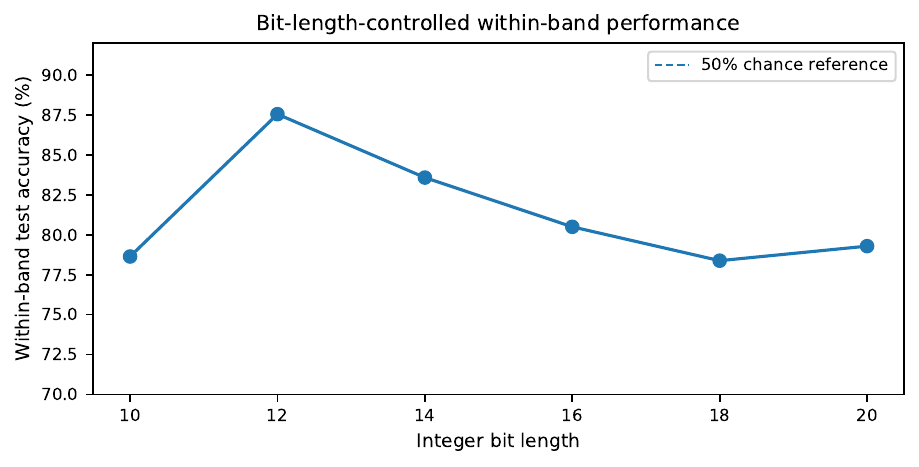}
\caption{Within-band test performance by integer bit length.}
\end{figure}

\subsection{Spatial reordering and cluster-count sensitivity}
\begin{figure}[H]
\centering
\includegraphics[width=0.68\linewidth]{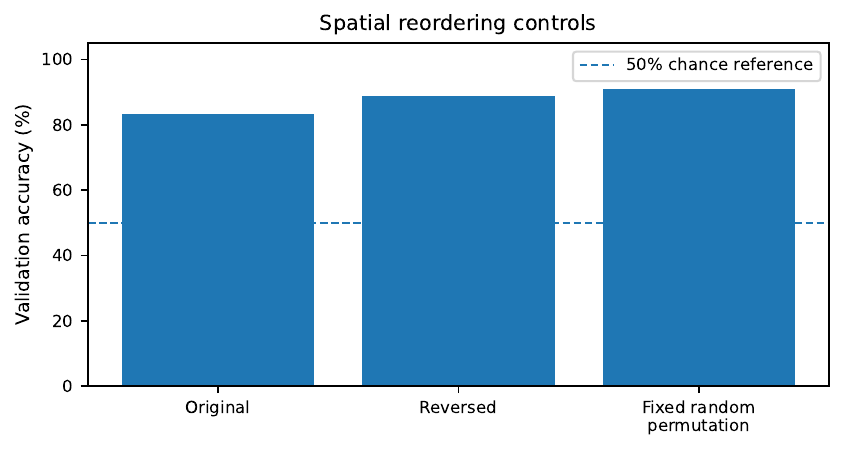}
\caption{Validation accuracy for original, reversed, and one fixed random bit permutation. All conditions preserve the same 32 bits.}
\end{figure}

\begin{figure}[H]
\centering
\includegraphics[width=0.68\linewidth]{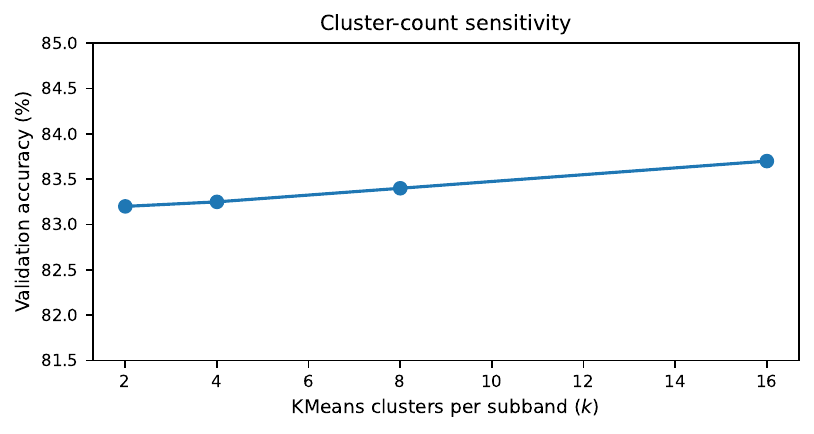}
\caption{Validation sensitivity to the number of k-means clusters per subband.}
\end{figure}

\end{document}